\pdfoutput=1

\documentclass[11pt]{article}

\usepackage[final]{acl}

\usepackage{times}
\usepackage{latexsym}

\usepackage[T1]{fontenc}

\usepackage[utf8]{inputenc}

\usepackage{microtype}

\usepackage{inconsolata}

\usepackage{graphicx}
\usepackage{times}
\usepackage{tabu}
\usepackage{latexsym}
\usepackage{stmaryrd}
\usepackage{amsmath}
\usepackage{multirow}
\usepackage{bbm}
\usepackage{graphicx}
\usepackage{amssymb}
\usepackage{booktabs}
\usepackage{algpseudocode}
\usepackage{algorithm}
\usepackage{graphicx}
\usepackage{caption}
\usepackage{subcaption}
\usepackage{tabularx}
\usepackage{colortbl}
\usepackage{xcolor}
\usepackage{tablefootnote}
\usepackage{enumitem}  
\usepackage{lipsum}
\usepackage{xspace}
\usepackage{bbding}
\usepackage{pifont}
\usepackage{arydshln}
\usepackage{array}
\usepackage{cleveref}
\newcommand{\PreserveBackslash}[1]{\let\temp=\\#1\let\\=\temp}
\newcolumntype{C}[1]{>{\PreserveBackslash\centering}p{#1}}
\newcolumntype{R}[1]{>{\PreserveBackslash\raggedleft}p{#1}}
\newcolumntype{L}[1]{>{\PreserveBackslash\raggedright}p{#1}}
\newcommand{\datasetname}{\texttt{IndoCareer}\xspace}

\definecolor{mygreen}{RGB}{217, 234, 211}
\definecolor{myred}{RGB}{244, 204, 204}

\newcommand{\ok}{\cellcolor{mygreen}}
\newcommand{\no}{\cellcolor{myred}}

\title{Cracking the Code: Multi-domain LLM Evaluation on \\Real-World Professional Exams in Indonesia}

\author{Fajri Koto \\
  Department of Natural Language Processing \\
  MBZUAI, Abu Dhabi, UAE \\
  \texttt{fajri.koto@mbzuai.ac.ae}}

\begin{document}
\maketitle
\begin{abstract}

While knowledge evaluation in large language models has predominantly focused on academic subjects like math and physics, these assessments often fail to capture the practical demands of real-world professions. In this paper, we introduce \datasetname{}, a dataset comprising 8,834 multiple-choice questions designed to evaluate performance in vocational and professional certification exams across various fields. With a focus on Indonesia, \datasetname{} provides rich local contexts, spanning six key sectors: (1) healthcare, (2) insurance and finance, (3) creative and design, (4) tourism and hospitality, (5) education and training, and (6) law. Our comprehensive evaluation of 27 large language models shows that these models struggle particularly in fields with strong local contexts, such as insurance and finance. Additionally, while using the entire dataset, shuffling answer options generally maintains consistent evaluation results across models, but it introduces instability specifically in the insurance and finance sectors.\footnote{Data can be accessed at \url{https://huggingface.co/datasets/indolem/IndoCareer}.}
\end{abstract}

\section{Introduction}

The evaluation of large language models (LLMs) has shifted from traditional natural language processing (NLP) tasks \cite{mikheev-etal-1999-named,straka-strakova-2017-tokenizing} to more complex, knowledge-intensive, and reasoning-based challenges. One of the key datasets used to assess these abilities is the massive multitask language understanding (MMLU) \cite{hendrycksmeasuring}. Initially introduced in English, MMLU datasets have also been developed in other languages, including Indonesian \cite{koto-etal-2023-large}, Chinese \cite{li-etal-2024-cmmlu}, and Arabic \cite{koto-etal-2024-arabicmmlu}. These datasets consist of school exam questions across various subjects and education levels, tailored to local curricula.\footnote{The English MMLU is based on the U.S. curriculum, while the Indonesian MMLU follows the Indonesian curriculum.} However, they primarily focus on academic subjects, often overlooking vocational and professional expertise, which are more relevant to real-world applications.

\begin{figure}[t]
    \centering
    \includegraphics[width=\linewidth]{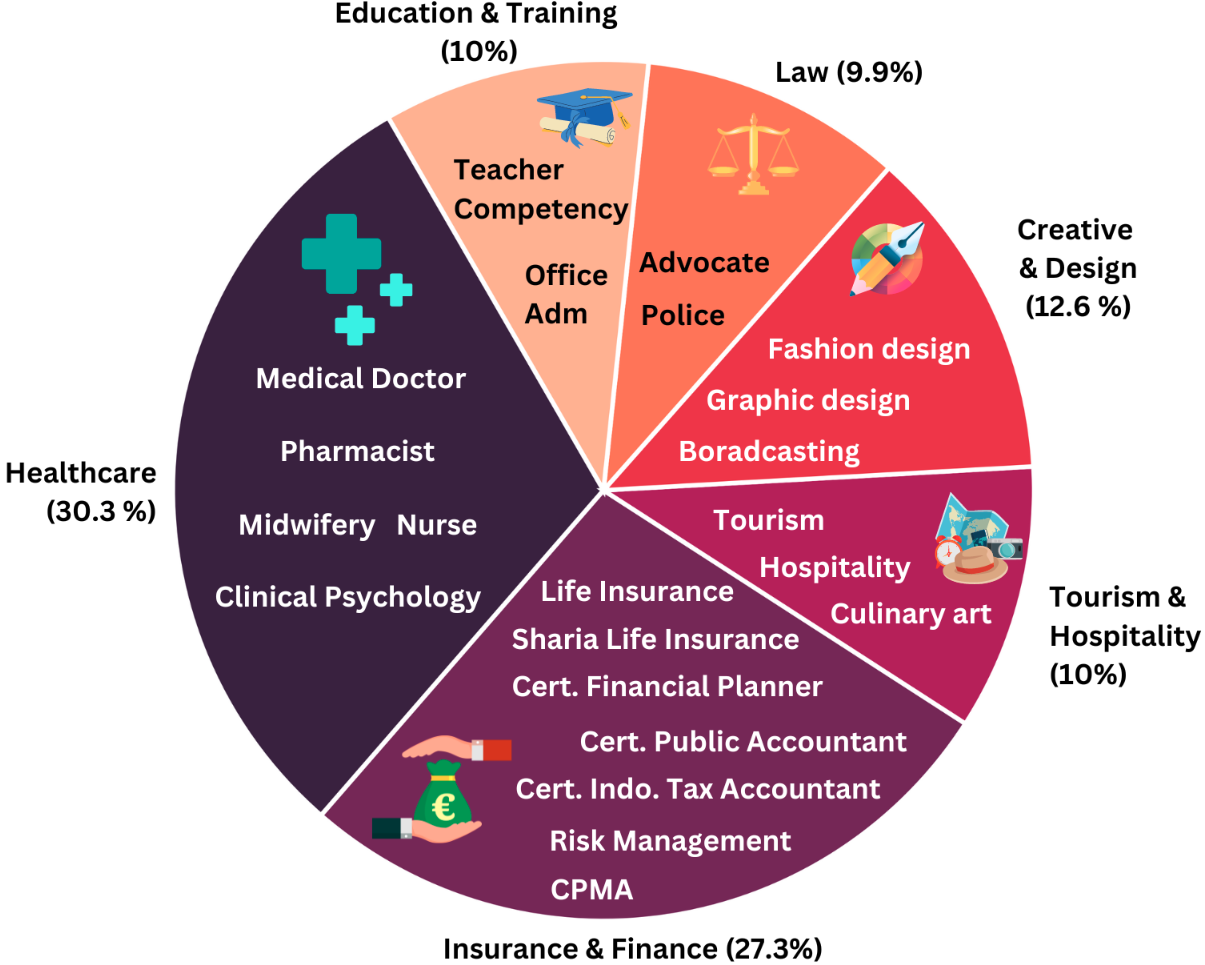} 
    \caption{Distribution of professions in \datasetname{}.}
    \label{fig:overview}
\end{figure}

Due to the recent widespread adoption of LLMs across various domains, including health \cite{zhang-etal-2024-llm-based}, education \cite{weijers-etal-2024-quantifying,srivatsa-kochmar-2024-makes}, and finance \cite{lee-soon-2024-finance}, evaluating a model's knowledge across professional fields has become crucial. For instance, in healthcare, the model must adhere to ethical standards \cite{gundersen2022future} and possess expertise in prevalent regional diseases. We should not trust AI-based health recommendations from models that have not passed a competency exam. Similarly, in education, the model needs to understand and align with local government teaching guidelines. Despite the importance of certification exams in professional fields, such exams have been largely excluded from prior work \cite{koto-etal-2023-large}.


In this paper, we introduce \datasetname{}, a dataset comprising 8,834 multiple-choice questions collected from various Indonesian competency exams, certification exams, and vocational school exams. Our focus on Indonesian addresses the limitations of prior work \cite{koto-etal-2023-large} and aims to enrich language diversity and local context nuances in NLP datasets, which are predominantly English-centric \cite{liu2024culturally}. Figure~\ref{fig:overview} shows the distribution of \datasetname{}, which covers 22 different professions across 6 categories: (1) healthcare, (2) insurance and finance, (3) creative and design, (4) tourism and hospitality, (5) education and training, and (6) law.  Additionally, we demonstrate that \datasetname{} is generally robust to option shuffling \cite{zhou-etal-2024-revisiting} when using the entire dataset, but it specifically introduces instability in insurance and finance professions.


\section{Related Work}

\paragraph{Indonesian Language Models} IndoBERT \cite{koto-etal-2020-indolem, wilie-etal-2020-indonlu}, IndoBERTweet \cite{koto-etal-2021-indobertweet},  IndoGPT \cite{cahyawijaya-etal-2021-indonlg}, and IndoBART \cite{cahyawijaya-etal-2021-indonlg} are among the earliest transformer-based language models developed from scratch for Indonesian. These models have been widely adopted by industry and academia across various applications. For models exceeding 1 billion parameters, no foundational models have been pre-trained exclusively on Indonesian text. Instead, research has focused on adapting multilingual models through fine-tuning techniques. Notable examples include Bactrian-X \cite{li2023bactrian}, which employs LoRA \cite{hu2022lora} for fine-tuning LLama-1 \cite{llama}, and Merak \cite{Merak}, Cendol \cite{cahyawijaya-etal-2024-cendol}, and Komodo \cite{komodo}, which are fine-tuned adaptations of LLama-2 \cite{touvron2023llama}. Despite growing interest in deploying Indonesian LLMs across various domains and job sectors, there remains a lack of suitable benchmarks tailored to evaluate their performance. To address this gap, we introduce \datasetname{}.

\paragraph{Benchmarks for Evaluating Language Models}
NusaCrowd \cite{cahyawijaya-etal-2023-nusacrowd} represents a significant effort to consolidate scattered datasets for Indonesian NLP. While most high-quality datasets focus on classical NLP tasks such as sentiment analysis, summarization, and text classification, benchmarks for knowledge-intensive and reasoning tasks have been notably limited until very recently. The introduction of IndoMMLU \cite{koto-etal-2023-large}, COPAL-ID \cite{wibowo-etal-2024-copal}, and IndoCulture \cite{koto2024indoculture} marks a step forward in this direction. COPAL-ID and IndoCulture focus on cultural commonsense reasoning, while IndoMMLU evaluates exam questions across different education levels in Indonesia, from primary to high school.

Despite recent advancements, a significant gap remains in evaluating LLMs on professional tasks in the Indonesian context, as IndoMMLU does not include questions from professional exams. This limitation is not unique to Indonesia; professional exam coverage is also limited in similar benchmarks for other languages. For example, English MMLU \cite{hendrycksmeasuring} and Chinese MMLU \cite{li-etal-2024-cmmlu} include professional exam questions in only 20\% of their datasets, while Arabic MMLU \cite{koto-etal-2024-arabicmmlu} has an even lower coverage of just 4\%. 

As LLMs are increasingly applied across various domains \cite{zhang-etal-2024-llm-based,lee-soon-2024-finance}, there is a pressing need for a benchmark that evaluates their readiness for professional job sectors. \datasetname{} addresses this gap, offering a comprehensive benchmark of professional exams spanning 22 professions, making it the first of its kind in Indonesia.

\section{IndoCareer}
\label{sec:data}

\begin{figure}[t]
    \centering
    \includegraphics[width=\linewidth]{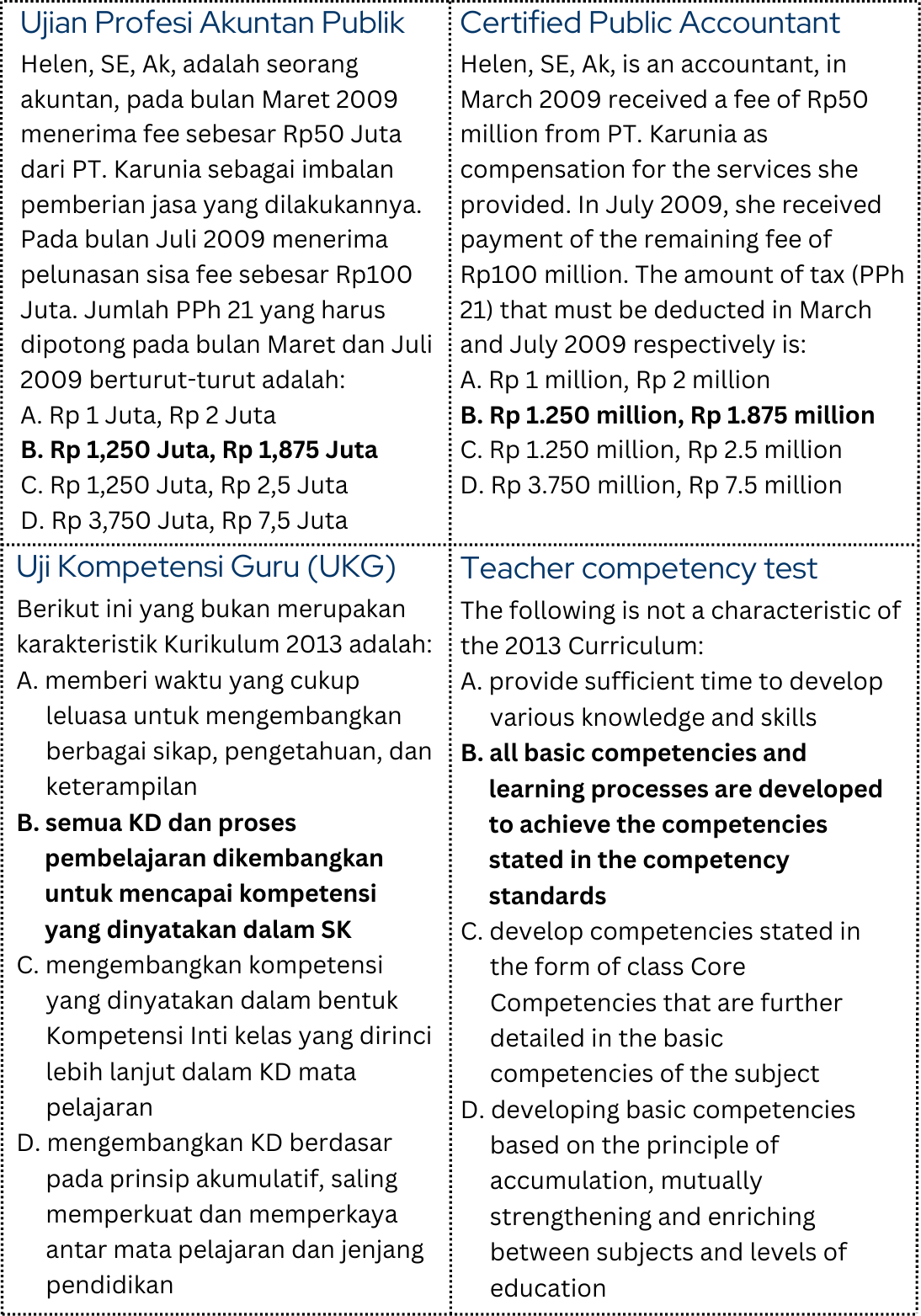} 
    \caption{Example of questions in \datasetname{}. The English translation is only for illustrative purposes.}
    \label{fig:example}
\end{figure}

\datasetname{} comprises 8,834 multiple-choice questions compiled from Indonesian competency exams, certification exams, and vocational school exams across 22 professions. In Indonesia, competency exams are commonly required in healthcare professions by the government. Certification exams, on the other hand, focus on specific skills within a profession, such as tax accounting in finance. At the high school level, vocational schools offer specialized training in areas like tourism, culinary arts, and fashion design.  In Figure~\ref{fig:overview}, Table~\ref{tab:subj_dist}, and Table~\ref{tab:stat}, we present detailed statistics for the 22 professions covered in \datasetname{}. In this dataset, we exclude engineering-related professions, as their certification exams are generally conducted in English.

\paragraph{Data Construction}  We manually collected exam questions from publicly available sources across 22 professions. A majority (78\%) of the questions were sourced from \texttt{Scribd},\footnote{\url{https://www.scribd.com/}} a document-sharing platform, while the remaining were obtained from local government websites\footnote{For example: \url{https://badanbahasa.kemdikbud.go.id}} and shared Google Drive folders. We ensured that all collected questions were relevant to their respective professions and suitable for distribution for research purposes. Importantly, 99\% of the exam questions were retrieved from file formats, such as PDFs and Word documents, rather than directly from web pages, minimizing the risk of overlap with training data used by LLMs.

To extract the questions and answers, we hired three professional teachers with Bachelor's degrees in Education for a one-month period. Their task focused exclusively on text-based questions, excluding any questions containing images (see Figure~\ref{fig:example} for examples). Each worker was responsible for extracting approximately 3,000 questions. To ensure ethical practices, they were compensated above the minimum wage in Indonesia, with the total workload equivalent to five full-time workdays.

\paragraph{Quality Control} We ensure the high quality of our dataset through a rigorous and multi-step quality control process. Although we employ ``expert'' workers who are native Indonesian speakers with at least a Bachelor's degree, additional measures are implemented to maintain and verify quality. First, all data sources are manually checked and validated by the author before being distributed to the workers. Workers also participate in a 1-hour workshop prior to data collection, ensuring they fully understand the guidelines and the expected data standards.

After the workers complete their tasks, we apply automated filtering to eliminate repetitive questions and entries without answer keys. To further validate the dataset, we conducted a manual review of 300 randomly selected samples (3.3\% of the dataset), performed by the authors of this paper. During this review, we verified the accuracy of the questions, answer options, and answer keys. The manual review achieved an accuracy rate of 99\%, demonstrating the dataset's reliability and representing the highest meaningfully achievable score for \datasetname{}.

\begin{table}[]
    \centering
    \resizebox{\linewidth}{!}{
        \begin{tabular}{lL{4cm}lc}
        \toprule
         \textbf{Field} &  \textbf{Professions} & \textbf{Exam Type} & \textbf{\#Q} \\        
        \midrule
            \multirow{5}{2cm}{Healthcare} & Medical Doctor &  Competency Exam & 805 \\
             &  Pharmacist&  Competency Exam& 598\\
             &  Midwifery & Competency Exam & 680\\
             &  Nurse& Competency Exam & 497\\
             &  Clinical Psychology& Other & 95\\
            \midrule
            \multirow{7}{2cm}{Insurance \& Finance} & Life Insurance & Certification Exam & 476\\
             & Sharia Life Insurance& Certification Exam & 558\\
             & CFP & Certification Exam & 96\\
             & CPA & Certification Exam & 663\\
             & CPMA & Certification Exam & 169\\
             & CITA & Certification Exam & 253\\
             & Risk Management & Certification Exam & 194\\
            \midrule
            \multirow{2}{2cm}{Tourism \& Hospitality} & Tourism &  Vocational School & 222\\
            &  Hospitality &  Vocational School & 367\\
            &  Culinary Art & Vocational School & 294\\
            \midrule
            \multirow{3}{2cm}{Creative \& Design} & Graphic Design &  Vocational School & 423\\
             &  Fashion Design & Vocational School & 267\\
             &  Broadcasting & Vocational School & 422\\
            \midrule
             \multirow{2}{2cm}{Law} & Advocate & Certification Exam & 591\\
            & Police  & Other & 280\\
            \midrule
             \multirow{2}{2cm}{Education \& Training} & Teacher Competency Test & Certification Exam & 538\\
            & Office Administration   & Vocational School & 346\\
        \bottomrule
        \end{tabular}
    }
    \caption{Number of questions in \datasetname{} across different professions. CFP stands for Certified Financial Planner, CPA stands for Certified Public Accountant, CPMA stands for Certified Professional Management Accountant, and CITA stands for Certified Indonesian Tax Accountant.}
    \label{tab:subj_dist}
\end{table}

\begin{table}[t]
    \centering
    \resizebox{\linewidth}{!}{
        \begin{tabular}{lrrr}
        \toprule
        \multirow{2}{*}{\textbf{Field}} & \multirow{2}{*}{\textbf{\# Questions}} & \multicolumn{2}{c}{\textbf{\# Chars}}\\
        \cmidrule{3-4}
        & & \textbf{Question} & \textbf{Answer} \\        
        \midrule
        Healthcare & 2675 & 277.3 & 95.9 \\ 
        Insurance and Finance & 2409 & 156.3 & 165.3 \\
        Tourism and Hospitality & 883 & 99.8 & 96.2 \\
        Creative and Design & 1112 & 101.0 & 100.5 \\
        Law & 871 & 130.7 & 141.2 \\
        Education and Training & 884 & 159.5 & 165.9 \\
        \bottomrule
        \end{tabular}
    }
    \caption{Average question and answer length (in characters) for each profession fields.}
    \label{tab:stat}
\end{table}

\begin{table*}[t]
    \centering
    \resizebox{\linewidth}{!}{
        \begin{tabular}{lC{2cm}C{2cm}C{2cm}C{1cm}C{2cm}C{2cm}C{1.5cm}}
        \toprule
       \multirow{2}{*}{\textbf{Model (\#parameters)}}& \multirow{2}{*}{\textbf{Healthcare}} & \textbf{Insurance} & \textbf{Tourism \&}  & \multirow{2}{*}{\textbf{Law}} &  \textbf{Creative} & \textbf{Education}& \multirow{2}{*}{\textbf{Average}} \\
       & & \textbf{\& Finance} & \textbf{Hospitality} & & \textbf{\& Design} & \textbf{\& Training} & \\
       \midrule
        Random & 20.6 & 25.8 & 20.0 & 24.1 & 20.1 & 22.8 & 22.5 \\
        \hdashline
        BLOOMZ (560M) & 17.9 & 23.9 & 19.3 & 27.5 & 17.6 & 24.9 & 21.3 \\
        BLOOMZ (1.7B) & 28.2 & 34.7 & 40.2 & 32.6 & 39.0 & 35.9 & 33.7 \\
        BLOOMZ (3B) & 29.8 & 39.2 & 42.2 & 37.3 & 44.2 & 40.8 & 37.3 \\
        BLOOMZ (7B) & 32.9 & 41.7 & 47.1 & 40.3 & 48.9 & 45.1 & 40.7 \\
        mT0$_\text{small}$ (300M) & 22.3 & 26.2 & 21.7 & 23.5 & 22.2 & 19.5 & 23.1 \\
        mT0$_\text{base}$ (580M) & 23.3 & 26.5 & 24.8 & 24.3 & 23.0 & 24.0 & 24.4 \\
        mT0$_\text{large}$ (1.2B) & 25.0 & 26.8 & 25.3 & 24.2 & 24.3 & 23.3 & 25.2 \\
        mT0$_\text{xl}$ (3.7B) & 27.7 & 38.9 & 43.8 & 36.0 & 42.4 & 43.3 & 36.6 \\
        mT0$_\text{xxl}$ (13B) & 29.4 & 41.1 & 44.3 & 40.0 & 46.1 & 44.1 & 38.7 \\
        Gemma-2 (2B) & 35.7 & 51.0 & 55.5 & 44.4 & 55.0 & 52.1 & 46.8 \\
        Gemma-2 (9B) & 54.3 & 62.2 & 68.0 & 56.9 & 68.1 & 60.8 & 60.5 \\
        Gemma-2 (27B) & 58.3 & 64.2 & 71.7 & 60.2 & 71.7 & 62.6 & 63.5 \\
        Aya-23 (8B) & 37.0 & 46.1 & 51.7 & 44.3 & 51.7 & 47.5 & 44.6 \\
        Aya-23 (35B) & 43.9 & 52.9 & 59.0 & 50.4 & 61.8 & 53.3 & 51.7 \\
        LLaMA-3.1 (8B) & 35.9 & 46.7 & 51.9 & 41.2 & 53.0 & 45.3 & 44.1 \\
        LLaMA-3.1$_\text{Instruct}$ (8B) & 44.8 & 53.6 & 61.1 & 47.7 & 63.3 & 54.9 & 52.4 \\
        LLaMA-3.1 (70B) & 61.4 & 65.0 & 69.4 & 64.0 & 72.3 & 61.4 & 64.8 \\
        LLaMA-3.1$_\text{Instruct}$ (70B) & \textbf{64.4} & \textbf{69.3} & \textbf{74.2} & \textbf{68.1} & \textbf{75.1} & \textbf{65.3} & \textbf{68.5} \\
        \hdashline
        Bactrian-ID (7B) & 20.5 & 29.0 & 22.7 & 26.6 & 25.5 & 25.1 & 24.7 \\
        IndoGPT (117M) & 21.5 & 26.6 & 24.5 & 23.2 & 18.1 & 23.6 & 23.2 \\
        Merak (7B) & 37.2 & 45.6 & 49.7 & 43.8 & 50.8 & 46.9 & 44.1 \\
        SeaLLM (7B) & \textbf{41.1} & \textbf{54.7} & \textbf{56.0} & \textbf{44.7} & \textbf{61.3} & \textbf{50.8} & \textbf{50.1} \\
        SEA-LION (7B) & 19.2 & 28.9 & 20.0 & 27.6 & 20.9 & 27.3 & 23.8 \\
        Komodo (7B) & 25.5 & 29.7 & 27.4 & 30.5 & 29.8 & 31.8 & 28.5 \\
        Cendol$_\text{mT5-xxl}$  (13B) & 20.8 & 24.8 & 22.9 & 22.9 & 21.8 & 21.4 & 22.5 \\
        Cendol$_\text{LLaMA2}$ (13B) & 23.3 & 28.6 & 22.7 & 24.7 & 24.0 & 25.2 & 25.1 \\
        \hdashline
        GPT-4o & \textbf{68.3} & \textbf{73.5} & \textbf{75.7} & \textbf{75.4} & \textbf{78.3} & \textbf{67.4} & \textbf{72.3} \\

        \bottomrule
        \end{tabular}
    }
    \caption{Zero-shot LLM performance (\% accuracy), combined across professional fields. ``Average'' means the average across all questions in \datasetname{}.}
    \label{tab:result}
\end{table*}

\paragraph{Data Statistics}  Table~\ref{tab:subj_dist} summarizes the distribution of questions in \datasetname{} across 22 professions, organized into six main fields: Healthcare, Insurance \& Finance, Tourism \& Hospitality, Creative \& Design, Law, and Education \& Training. Each profession corresponds to specific exam types, including competency exams, certification exams, vocational school exams, and others. Healthcare encompasses five professions, such as Medical Doctor and Pharmacist, contributing a total of 2,675 questions. Insurance \& Finance, the largest category with seven professions, includes fields like Life Insurance, Certified Public Accountant (CPA), and Risk Management, with 2,409 questions. Tourism \& Hospitality covers three professions—Tourism, Hospitality, and Culinary Art—comprising 883 questions, while Creative \& Design features 1,112 questions. The Law field includes Advocate and Police exams, with a total of 871 questions, while Education \& Training, with Teacher Competency Tests and Office Administration, adds another 884 questions. 

According to Table~\ref{tab:stat}, healthcare questions are the longest, averaging 2 to 3 times the length of those in tourism and hospitality, and creative and design. The number of multiple-choice options is generally consistent across professional fields, averaging 4 options. However, the total character count of the options varies, with insurance and finance, and education and training having the longest options, exceeding 160 characters. 

Additionally, we manually examined 300 random samples to assess whether answering the questions required local context.\footnote{The 300 random samples are the same as those used for the manual review. Given the 99\% accuracy rate from the initial review, we included an additional 1\% of randomly selected correct samples for the local context assessment.}  Our analysis revealed that 34\% of the questions incorporated Indonesian local context, with a notable concentration in the fields of insurance and finance, tourism and hospitality, and law.



\section{Experiments}

\citet{pezeshkpour-hruschka-2024-large,zhou-etal-2024-revisiting} demonstrated that LLMs are highly sensitive to the order of options in multiple-choice questions. To ensure a more robust evaluation, we report the average performance across three evaluations for each model: one using the original order of options and two with the options shuffled.\footnote{For reproducibility, we also release two versions of \datasetname{} with shuffled options, available at \url{https://huggingface.co/datasets/indolem/IndoCareer}.} We evaluated one closed-source model (GPT-4o) and 26 open-weight LLMs, comprising 18 multilingual models (BLOOMZ \cite{muennighoff2022crosslingual}, mT0 \cite{muennighoff2022crosslingual}, Gemma-2 \cite{team2024gemma}, Aya-23 \cite{ustun-etal-2024-aya}, LLaMA3.1\footnote{\url{https://github.com/meta-llama/llama3}}) and 8 Indonesian-centric models (IndoGPT \cite{cahyawijaya-etal-2021-indonlg}, Bactrian-ID \cite{li2023bactrian}, Merak \cite{Merak}, Komodo \cite{komodo}, SeaLLM \cite{nguyen2023seallms}, SEA-LION \cite{sea_lion_2023}, and Cendol \cite{cahyawijaya-etal-2024-cendol}). Details for each model can be found in the Appendix.

Our focus is on zero-shot experiments using the Indonesian prompt: \textit{Ini adalah soal [subject] untuk [exam type]. Pilihlah salah satu jawaban yang dianggap benar!}.\footnote{The English translation is "This is a [subject] question for [exam type]. Please choose the correct answer!"} 
For evaluation, we use the \texttt{LM-Harness} package \cite{eval-harness}, selecting the answer based on the highest probability of the first token (i.e., A, B, C, D) in the generated output. Specifically, for GPT-4o, we used the \texttt{gpt-4o} model from OpenAI,\footnote{\url{https://openai.com/}} selecting the answer based on the first letter generated in the output.\footnote{For GPT-4o, we slightly adjusted the prompt, instructing the model to output only one of the options as the answer.} 

\subsection{Results}

Table~\ref{tab:result} summarizes the zero-shot performance of various large language models (LLMs) across professional fields in \datasetname{}, highlighting significant differences in their ability to handle Indonesian professional exams. GPT-4o and LLaMA-3.1 (70B) emerge as the top-performing models, with GPT-4o achieving the highest overall accuracy at 72.3\%, followed closely by LLaMA-3.1 (70B) with 68.5\%. This 4-point gap demonstrates GPT-4o's superior capability in handling complex tasks across diverse professions. In contrast, other multilingual models show significantly lower accuracy, ranging between 38.0\% and 60.0\%, indicating their struggles with Indonesian-specific professional exams.

Indonesian-centric models, including SEA-LION, Komodo, and Cendol, underperform dramatically, with results close to random guessing in some fields. These findings suggest that existing Indonesian-centric models are not yet optimized for professional exam tasks, limiting their utility in practical applications. Notably, the SEA-LION (7B) and Komodo (7B) models achieve only 23.8\% and 28.5\% average accuracy, respectively, underscoring the gap between local adaptations and the more capable multilingual models.

Healthcare stands out as the most challenging professional field, with an average performance across all models at only 37.2\%.\footnote{This figure is calculated by averaging all values in the Healthcare column of Table~\ref{tab:result}.} This poor performance underscores the limitations of current off-the-shelf LLMs as reliable health advisors in the Indonesian context. These findings highlight the critical need for robust model adaptations and fine-tuning specifically tailored to Indonesian professional tasks to enhance performance and to ensure applicability in high-stakes domains such as healthcare.

\subsection{Analysis}

\paragraph{Shuffling the multiple-choice options leads to unstable results in insurance and finance.} Table~\ref{tab:std} lists the top 10 professions with the highest standard deviation ($\sigma$) in performance across three evaluation runs. While the standard deviations are relatively low, ranging from 1.5 to 3.0, they indicate minor instabilities in model predictions when the multiple-choice options are shuffled. For certain professions, such as Certified Financial Planner, Certified Indonesian Tax Accountant, and Certified Professional Management Accountant, the average rank correlation ($\tau$) drops below 0.9, indicating reduced consistency in model performance across evaluation runs. Although their deviations are not severe, they highlight areas where models are less robust to option shuffling, particularly in domains requiring nuanced reasoning. Across the entire dataset, however, the rank correlation remains high, with an average of 0.98. This indicates that while minor instabilities exist at the profession level, the overall dataset maintains stable performance.


\begin{table}[t]
    \centering
    \resizebox{\linewidth}{!}{
        \begin{tabular}{lrr}
        \toprule
        \textbf{Profession} & {$\sigma \downarrow$} & \textbf{$\tau \uparrow$} \\
        \midrule
        Clinical Psychology & \no 3.00 & 0.93 \\
        Cert. Financial Planner & \no 2.91 & \no 0.68 \\
        Cert. Professional Management Accountant  & \no 2.00 & \no 0.90 \\
        Fashion Design & 1.98 & 0.93 \\
        Advocate & 1.96 & 0.91 \\
        Police & 1.86 & 0.95 \\
        Cert. Indo. Tax Accountant & 1.81 & \no 0.85 \\
        Sharia Life Insurance & 1.81 & 0.97 \\
        Risk Management & 1.76 & 0.97 \\
        Tourism & 1.63 & 0.96 \\
    \midrule
    All & 1.57 & 0.98 \\
        \bottomrule
        \end{tabular}
    }
    \caption{Top 10 professions with the highest standard deviation ($\sigma$). $\tau$ represents the average rank correlation across three runs. The red cells are the three worse score. The scores are based on evaluations across 27 models. }
    \label{tab:std}\end{table}
 
\begin{figure}[t]
    \centering
    \includegraphics[width=\linewidth]{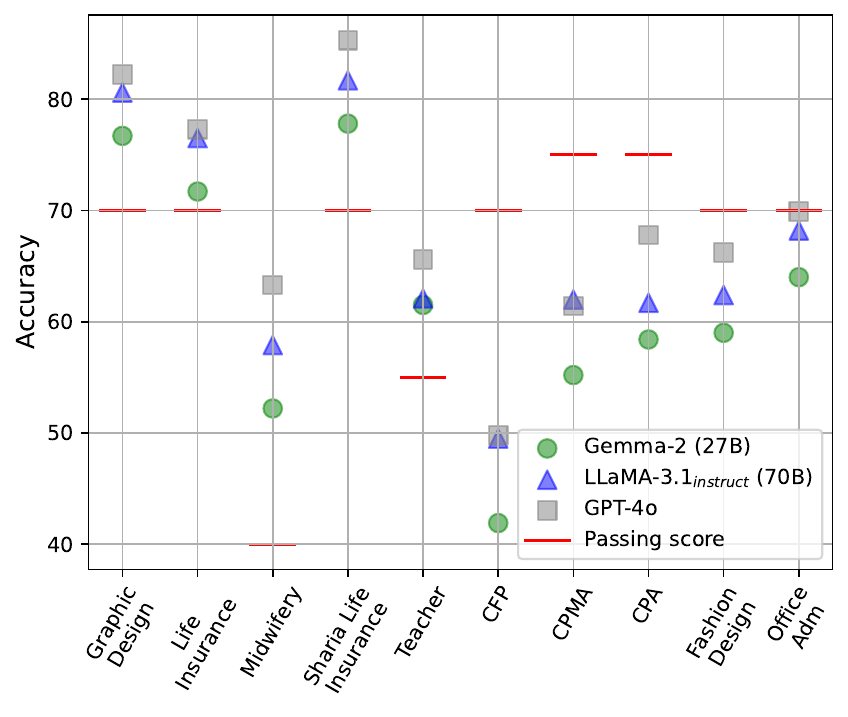} 
    \caption{Top 5 and bottom 5 professions based on the model's accuracy disparity relative to the passing score.}
    \label{fig:analysis}.
\end{figure}

\paragraph{LLMs perform well in life insurance certification but struggle with finance-related certifications.}
Figure~\ref{fig:analysis} illustrates the performance of LLMs across the top 5 and bottom 5 professions in terms of accuracy relative to the passing scores. The passing scores for each exam, represented by red horizontal lines, were sourced from publicly available information. The figure highlights that while GPT-4o, LLaMA-3.1 (70B), and Gemma-2 (27B) achieve passing scores for professions such as life insurance, sharia life insurance, graphic design, midwifery, and teacher competency, they fall significantly short for finance-related certifications.

None of the models evaluated pass the exams for Certified Financial Planner (CFP), Certified Professional Management Accountant (CPMA), Certified Public Accountant (CPA), fashion design, or office administration. Notably, GPT-4o, the best-performing model overall, falls over 20 points below the passing score for CFP, emphasizing the difficulty of finance-related tasks.  The results suggest that finance-related certifications, which often require domain-specific reasoning and detailed calculations, remain a challenge for current LLMs. On the other hand, professions with more straightforward knowledge requirements, such as life insurance or midwifery, align better with the strengths of existing LLMs. These findings highlight the need for targeted fine-tuning and adaptation to improve performance in specialized and calculation-heavy fields like finance.


\paragraph{Questions with local context and numerical analysis pose greater challenges.} 
We conducted an error analysis on the best-performing open-weight model, LLaMA-3.1 (70B), by examining 100 incorrectly predicted samples and 100 correctly predicted samples for comparison. These samples were drawn from the original questions, without applying option shuffling. The analysis showed that questions with Indonesian local context were more common among the incorrectly predicted samples, with 50\% of the incorrect predictions containing local context, compared to only 22\% among the correct predictions. Considering that \datasetname{} contains 34\% local context overall, as discussed in Section~\ref{sec:data}, this suggests that questions incorporating local context are particularly challenging for language models. This finding aligns with prior research \cite{koto2024indoculture}, indicating that questions grounded in local context often introduce cultural or situational nuances not well-captured in the models' pretraining data.

In addition to local context, questions involving numerical analysis also posed significant challenges for LLaMA-3.1 (70B). Among the incorrectly predicted samples, 43 required numerical reasoning, compared to only 29 among the correctly predicted ones. Numerical questions often involve calculations or logical reasoning steps, which many LLMs are not explicitly optimized to handle. These results reveal two key areas where model performance could be improved: understanding and addressing culturally specific content and enhancing their capabilities for numerical reasoning. 


\section{Conclusion}

We introduce \datasetname{} as the most comprehensive dataset of professional exams across various job sectors in Indonesia. The dataset encompasses 22 professions, categorized into healthcare, insurance and finance, creative and design, tourism and hospitality, education and training, and law. Evaluations across different LLMs show that most off-the-shelf models demonstrate vocational and professional expertise below the passing scores. We believe \datasetname{} will be valuable in supporting LLM adaptation for various job sectors in Indonesia.

\section*{Limitations}

There are three main limitations to our work: (1) \datasetname{} excludes multimodal data such as tables, audio, images, and videos. Including these would make the benchmark more comprehensive and reflective of real-world scenarios. However, since our focus is on LLM evaluation, we only include text-based questions; (2) Engineering-related professions are excluded from \datasetname{} because the language used in these exams is primarily English, while our focus is on the Indonesian language; (3) The evaluation is limited to multiple-choice questions and does not include text generation tasks. We follow prior work in using the multiple-choice format as an initial step to address the lack of professional and vocational exam benchmarks in Indonesian.

\section*{Ethical Considerations}

\datasetname{} is released under the Creative Commons Attribution-NonCommercial-ShareAlike 4.0 International License\footnote{\url{https://creativecommons.org/licenses/by-nc-sa/4.0/}} and is intended solely for academic research. The questions included in \datasetname{} are sourced from publicly available materials. We collected these questions in compliance with Indonesian Copyright Law No. 28 of 2014, specifically Article 44. This article states that the use, reproduction, and/or modification of works or related rights, in whole or in part, is not considered copyright infringement, provided the source is properly cited and the purpose is for education or research.\footnote{\url{https://wipolex-res.wipo.int/edocs/lexdocs/laws/en/id/id064en.pdf}}

\bibliography{custom}

\newpage
\appendix
\vspace{0.5cm}
\section{Models}

\begin{table}[h!]
    \centering
    \resizebox{\linewidth}{!}{
        \begin{tabular}{lr}
        \toprule
        \textbf{Models (\#parameters)} & \textbf{Source} \\
        \midrule
        BLOOMZ (560M)  & \texttt{bigscience/bloomz-560m} \\
        BLOOMZ (1.1B)  & \texttt{bigscience/bloomz-1b1} \\
        BLOOMZ (1.7B)  & \texttt{bigscience/bloomz-1b7} \\
        BLOOMZ (3B)  & \texttt{bigscience/bloomz-3b} \\
        BLOOMZ (7.1B)  & \texttt{bigscience/bloomz-7b1} \\
        \hdashline 
        mT0$_\text{small}$ (300M)   & \texttt{bigscience/mt0-small} \\
        mT0$_\text{base}$ (580M)  & \texttt{bigscience/mt0-base} \\
        mT0$_\text{large}$ (1.2B)   & \texttt{bigscience/mt0-large} \\
        mT0$_\text{xl}$ (3.7B)   & \texttt{bigscience/mt0-xl} \\
        mT0$_\text{xxl}$ (13B)  & \texttt{bigscience/mt0-xxl} \\
        \hdashline
        Gemma-2 (2B) & \texttt{google/gemma-2-2b-it} \\
        Gemma-2 (9B)& \texttt{google/gemma-2-9b-it} \\
        Gemma-2 (27B) & \texttt{google/gemma-2-27b-it} \\
        \hdashline 
        Aya-23 (8B) &  \texttt{CohereForAI/aya-23-8B} \\
        Aya-23 (35B) & \texttt{CohereForAI/aya-23-35B} \\
        \hdashline 
        LLaMA3.1 (8B)  & \texttt{meta-llama/Meta-Llama-3.1-8B} \\
        LLaMA3.1-Instruct (8B)  & \texttt{meta-llama/Meta-Llama-3.1-8B-Instruct} \\
        LLaMA3.1 (70B)  & \texttt{meta-llama/Meta-Llama-3.1-70B} \\
        LLaMA3.1-chat (70B)  & \texttt{meta-llama/Meta-Llama-3.1-70B-Instruct} \\
        \hdashline 
        Bactrian-ID (7B) & \texttt{haonan-li/bactrian-id-llama-7b-lora} \\
        IndoBART (132M)  & \texttt{indobenchmark/indobart-v2} \\
        IndoGPT (117M)  & \texttt{indobenchmark/indogpt} \\
        Merak (7B)  & \texttt{Ichsan2895/Merak-7B-v5-PROTOTYPE1} \\
        SeaLLM (7B)  & \texttt{SeaLLMs/SeaLLMs-v3-7B-Chat} \\
        SEA-LION (7B)  & \texttt{aisingapore/sea-lion-7b} \\
        Komodo (7B)  & \texttt{Yellow-AI-NLP/komodo-7b-base} \\
        Cendol$_\text{mT5-xxl}$ (13B) & \texttt{indonlp/cendol-mt5-xxl-merged-inst}\\
        Cendol$_\text{LLaMA2}$ (13B) & \texttt{indonlp/cendol-llama2-13b-merged-chat} \\
        \bottomrule
        \end{tabular}
    }
    \caption{With the exception of GPT-4o, all the models used in this study were sourced from Huggingface \cite{wolf-etal-2020-transformers}.}
    \label{tab:models}
\end{table}

\section{Full Results}
Table~\ref{tab:detail_result} presents the accuracy of each model across various professions. The passing scores for each exam were sourced from publicly available information. We found that GPT-4o passes most of the exams, with the exceptions being Certified Financial Planner, Certified Public Accountant, Certified Professional Management Accountant (CPMA), and Office Administration. LLaMA-3.1 and Gemma-2 also pass some Indonesian exams, but no Indonesian-centric model has yet passed the professional and vocational exams in Indonesia.



\newpage

\begin{table*}[t]
    \centering
    \resizebox{\linewidth}{!}{
        \begin{tabular}{lcccccccccccc}
        \toprule
       \textbf{Profession} & \textbf{P.Score} & \textbf{BLOOMZ} & \textbf{mT0} & \textbf{Aya-23} & \textbf{Gemma-2} & \textbf{LLaMA3.1} & \textbf{Merak}  & \textbf{SeaLLM} & \textbf{SEA-LION} & \textbf{Komodo} & \textbf{Cendol} & \textbf{GPT-4o} \\
       \midrule
        \multicolumn{12}{l}{\textbf{Healthcare}}  \\
        Medical Doctor & 66.0 & 33.4 & 27.3 & 45.1 & 61.6 & \ok 70.9 & 39.6 & 42.9 & 19.9 & 24.1 & 23.1 & \ok 74.8 \\
        Pharmacist & 57.0 & 30.0 & 24.4 & 41.2 & 55.8 & \ok 62.9 & 36.2 & 40.3 & 19.4 & 23.0 & 23.0 & \ok 64.5 \\
        Midwifery & 40.0 & 29.4 & 30.9 & 42.2 & 52.2 & \ok 57.9 & 33.1 & 37.4 & 20.5 & 22.8 & 22.2 & \ok 63.3 \\
        Nurse & 60.0 & 35.3 & 33.9 & 45.5 & 58.4 &\ok  60.8 & 36.2 & 44.7 & 23.8 & 25.6 & 25.7 & \ok 66.4 \\
        Clinical Psycology & 70.0 & 53.3 & 50.7 & 62.7 & \ok 73.6 & 71.4 & 53.3 & 62.7 & 26.4 & 33.7 & 34.1 & \ok 74.3 \\
        \midrule
        \multicolumn{12}{l}{\textbf{Insurance \& Finance}}  \\
        Life Insurance & 70.0 & 48.7 & 48.1 & 61.3 & \ok 71.7 & \ok 76.5 & 49.0 & 59.3 & 27.3 & 35.0 & 30.2 & \ok 77.3 \\
        Sharia Life Insurance & 70.0 & 45.9 & 47.1 & 61.6 & \ok 77.8 & 81.7 & 51.7 & 64.1 & 32.2 & 30.0 & 31.8 & \ok 85.3 \\
        Cert. Financial Planner & 70.0 & 29.4 & 27.6 & 36.9 & 41.9 & 49.5 & 25.4 & 38.0 & 24.0 & 22.6 & 22.2 & 49.8 \\
        Cert. Public Accountant & 75.0 & 37.9 & 36.7 & 44.7 & 58.4 & 61.7 & 42.8 & 48.2 & 25.9 & 27.0 & 26.6 & 67.8 \\
        Cert. Indo. Tax Accountant & 60.0 & 37.5 & 39.1 & 41.7 & 47.3 & 50.4 & 34.3 & 45.5 & 32.3 & 33.2 & 31.5 & \ok 60.7 \\
        CPMA & 75.0 & 32.7 & 28.7 & 40.6 & 55.2 & 62.0 & 38.6 & 40.8 & 26.9 & 25.9 & 24.1 & 61.4 \\
        Risk Management & 70.0 & 41.7 & 37.9 & 51.1 & 64.0 & 67.2 & 45.0 & 53.2 & 26.9 & 29.3 & 32.1 & \ok 70.9 \\
        \midrule
        \multicolumn{12}{l}{\textbf{Tourism \& Hospitality}}  \\
        Tourism & 70.0 & 51.3 & 53.6 & 58.1 & 72.6 & \ok 74.3 & 45.8 & 58.8 & 21.8 & 30.1 & 24.2 & \ok 76.6 \\
        Hospitality & 70.0 & 43.0 & 43.4 & 54.8 & 67.2 & 69.0 & 47.6 & 55.4 & 23.4 & 25.1 & 26.4 & \ok 71.7 \\
        Culinary Art & 70.0 & 47.0 & 45.2 & 62.0 & 73.5 & \ok 76.7 & 50.1 & 60.4 & 22.8 & 28.3 & 22.2 & \ok 79.5 \\
        \midrule
        \multicolumn{12}{l}{\textbf{Creative \& Design}}  \\
        Fashion Design & 70.0 & 34.8 & 35.2 & 47.1 & 59.0 & 62.4 & 36.4 & 49.1 & 20.6 & 25.3 & 21.7 & 66.2 \\
        Graphic Design & 70.0 & 52.9 & 53.8 & 65.3 & \ok 76.7 & 80.6 & 54.8 & 65.0 & 23.6 & 29.3 & 28.0 & \ok 82.2 \\
        Broadcasting & 70.0 & 51.6 & 49.2 & 63.9 & \ok 75.3 & \ok 77.3 & 54.7 & 63.9 & 23.4 & 30.7 & 25.0 & \ok 79.8 \\
        \midrule
        \multicolumn{12}{l}{\textbf{Law}}  \\
        Advocate & 70.0 & 34.6 & 39.9 & 47.1 & 59.9 & 68.7 & 36.7 & 41.9 & 26.4 & 27.3 & 26.4 & \ok 72.9 \\ 
        Police & 60.0 & 44.2 & 37.4 & 47.7 & 56.2 & \ok 64.0 & 45.4 & 47.5 & 21.7 & 27.0 & 26.5 & \ok 67.6 \\
        \midrule
        \multicolumn{12}{l}{\textbf{Education \& Training}}  \\
        Teacher Competency & 55.0 & 46.6 & 44.1 & 53.4 & \ok 61.5 & \ok 62.1 & 45.5 & 48.8 & 27.5 & 29.3 & 27.7 & \ok 65.6 \\ 
        Office Administration & 70.0 & 38.9 & 42.1 & 52.1 & 64.0 & 68.2 & 42.0 & 50.0 & 24.2 & 27.7 & 23.0 & 69.9 \\
        \bottomrule
        \end{tabular}
    }
    \caption{Zero-shot LLM performance (\% accuracy) across professions for each model. ``P.Score'' indicates the passing score for each exam. The models used in this table include BLOOMZ (7B), mT0$_\text{xxl}$, Aya-23 (35B), Gemma-2 (27B), LLaMA-3.1$_\text{Instruct}$, Merak (7B), SeaLLM (7B), SEA-LION (7B), Komodo (7B), Cendol$_\text{LLaMA2}$ (13B) and GPT-4o. Green cells indicate that the model meets or exceeds the passing score.}
    \label{tab:detail_result}
\end{table*}

\end{document}